# Detecting and explaining clinical-omics inconsistencies to improve patient cohort stratification: an application to Parkinson's disease

José A. Pardo-Pérez[a]; Jaime Ñíguez[a]; Ana Luisa Gil-Martínez[a,b]; Tomás Bernal[a]; Guillermo Rocamora[c,d]; Mina Ryten[c,d]; Laura Ibáñez[e]; José T. Palma[a]; Juan A. Botía[a,f*]; Alicia Gómez-Pascual[a*]

[a]Department of Information and Communications Engineering, Faculty of Computer Science, University of Murcia, Murcia, Spain.
[b]Department of Clinical and Movement Neurosciences, UCL Queen Square Institute of Neurology, London, UK.
[c]UK Dementia Research Institute, University of Cambridge, Cambridge, United Kingdom.
[d]Department of Clinical Neurosciences, School of Clinical Medicine, University of Cambridge, Cambridge, United Kingdom.
[e]Department of Neurology, Washington University School of Medicine, Saint Louis, MO, 63110, USA.
[f]Department of Neurodegenerative Disease, UCL Queen Square Institute of Neurology, London, UK.

**Corresponding authors information**
*Juan A. Botía - juanbot@um.es - +34868887317

- Total number of words of the manuscript: 6692
- Number of words of the abstract: 200
- Number of figures: 2
- Number of tables: 0

# Abstract

Discrepancies between clinical diagnoses and omics profiles within a characterized cohort may reflect misdiagnosis, hidden subgroups or prodromal disease states. We propose MLASDO, a tool to detect and characterize such discrepancies before downstream analyses. MLASDO (1) detects outliers using two unsupervised methods, thereby flagging potential poor-quality samples; and (2) identifies and characterizes anomalous samples (ASs), i.e., individuals whose molecular profile resembles the opposite clinical class. We applied MLASDO to the Parkinson's Progression Markers Initiative (PPMI) and Parkinson's Disease Biomarkers Program (PDBP) Parkinson's disease (PD) cohorts. In PPMI, it detected 26 outliers and 12 ASs: 5 anomalous healthy controls (AHCs) and 7 anomalous PD cases (APDs). AHCs exhibited higher cerebrospinal fluid (CSF) $A\beta_{1\text{–}42}$ levels than controls ($P < 0.0396$), suggesting resistance to cognitive decline. AHC-specific genes were enriched for the *MAPK* pathway ($P$<0.0145), implicated in PD pathogenesis. One AHC later received a different neurological disorder, three months after enrollment. In PDBP, it identified 10 outliers and 8 ASs: 2 AHCs and 6 APDs. One AHC exhibited 24 clinical features consistent with a PD-like phenotype, including severe motor impairment. Results are compiled in an interactive report for inspection and querying, highlighting clinically meaningful individuals otherwise overlooked in conventional analyses.

## 1. Introduction

Complex diseases are caused by a combination of genetic, lifestyle, and environmental factors [1] and characterized by a high heterogeneity in terms of age at onset and clinical manifestations [2,3]. This significantly complicates early diagnosis and accurate prognosis assessment. Studies based on case/control cohort (CCC) individuals with standardized clinical descriptions and available omics samples are an important resource for research. And we need methods for leveraging the abundance and complexity of the heterogeneous sources of data they offer. We hypothesize that omics data can help better shape patient cohorts and refine patient classification by uncovering discrepancies between clinical diagnoses and molecular phenotypes within cohorts. Our focus is on CCCs, with clinical and omics data for all individuals in the context of a specific disease [4–9]. To what extent do all diagnosed individuals share similar omics profiles? Is there any correspondence between normal and disease individuals as defined by clinical data and groups of the same individuals that naturally emerge from omics? Can we leverage the availability of clinical and omics data to identify clinical-omics inconsistencies and use them in better shaping the cohort? Plausible answers to all these questions pass through the detection of high-quality patient strata and the characterization of concordance levels between clinical and omics profiles. The first step towards this is to characterize outliers and anomalous samples (ASs). Outliers are samples that deviate from the main sample data distribution of the cohort and may indicate technical artifacts or quality-related issues. ASs are those whose omics profile more closely resembles that of the opposite clinical class (e.g., a healthy control that comes with a disease-like omic profile).

To detect omics-based outliers in CCC datasets, several unsupervised algorithms have been applied, including: (1) density-based methods such as Local Outlier Factor [10], which scores individuals by comparing their local density with that of their nearest neighbors; (2) tree-based methods such as Isolation Forest [11], which isolates outlier samples through random recursive partitioning, exploiting the fact that these samples require fewer splits to be separated from the rest; (3) kernel-based methods such as One-Class Support Vector Machine (SVM) [12], which learns a boundary in a kernel-induced feature space that separates individuals and (4) low-rank decomposition methods such as Robust Principal Component Analysis [13], which captures the structure of outliers in a sparse, small matrix as a result of the factorization. Given their unsupervised character, these methods are primarily informative about technical quality and are commonly used to flag low-quality samples that may be removed prior to downstream analyses.

In contrast, the detection of ASs can be seen as a supervised classification problem because ASs are data points not correctly handled by the classifier model based on the omics, using the individual's diagnostics as the target variable. We can find in the literature approaches to detect ASs like: (1) Regression-based methods such as sparse label-noise-robust logistic regression (Rlogreg) [14]; (2) SVM-based approaches such as recursive SVM [15] and SVM formulations with ramp loss [16,17], where the misclassified large-margin samples are identified as ASs and feature selection can be incorporated by limiting the number of support vectors and (3) Tree-based proximity methods such as random forest geometry and accuracy-preserving (RF-GAP) proximities [18]. Despite the wide range of algorithms available for detecting outliers and ASs, to the best of our knowledge, no available software makes distinctions between them. For outlier detection, general-purpose libraries such as PyOD [19] have been developed, alongside domain-specific tools including bulk RNA-seq tools like EnsMOD [20], and single-cell approaches such as SCISSORS [21]. On the other hand, ROSIE [22] identifies ASs through a robust sparse ensemble framework.

In this paper, we propose to go beyond simply detecting samples of interest. To this end, we developed MLASDO (Machine Learning based Anomalous Sample Detection on Omics), a new software that identifies outliers and ASs and deeply annotates ASs with clinically and molecularly relevant information in CCC datasets to be used for decision-making during downstream analyses. Its workflow comprises four main steps: (1) detection of both outliers and ASs; (2) clinical (or metadata-based) characterisation of ASs; (3) molecular characterization of ASs; and (4) automated highly detailed report generation for further decision-making. Outliers are detected using two unsupervised algorithms and inductive biases to provide two alternative pieces of evidence: Isolation Forest [11] and robust PCA based on the Hubert estimator [23]. ASs are identified using an SVM classifier [24] trained to classify clinical groups based on the available transcriptomics. It leverages the inductive bias of the SVM algorithm by using, as a measure of anomaly, the distance of each transcriptomic sample to the SVM hyperplane. After AS identification, our tool clinically annotates and explains each AS identified by comparing its clinical data to what is expected in its diagnostic group or origin. It reports on control individuals showing a disease-like transcriptomic signature along with the covariates that may explain this anomaly. Analogously, for disease individuals with a healthy-like transcriptomic signature. Besides, it reveals the genes and pathways altered in each individual, using SHapley Additive exPlanations (SHAP) [25,26].

To showcase MLASDO, we used it on whole-blood transcriptomics data from the Parkinson's Progression Markers Initiative (PPMI) [4], which includes a total of 164 healthy controls (HC) and 309 idiopathic Parkinson's disease (PD) cases. In this cohort, the tool detected a total of 26 outliers and 12 ASs: five anomalous HC (AHC) and seven anomalous PD cases (APD) based on their transcriptomic signature. As a secondary additional example, we apply the tool to the PDBP (Parkinson's Disease Biomarkers Program) cohort [5], whose details appear in the Online Resource (see below). The reports generated by the tool for these cohorts are at the tool's GitHub repository (https://github.com/joseadrianpp/MLASDO). In summary, the tool can support clinical research by improving cohort characterization and stratification, and by identifying individuals with atypical molecular or clinical profiles who may merit closer follow-up or further investigation.

# 2. Methods

MLASDO requires an omic dataset, a clinical dataset, and an optional domain expert knowledge table (DEKT). The DEKT is a user-provided table that specifies, for each clinical covariate, the threshold and the direction of change that associate an individual with the case (e.g., PD) or the control group and is used to annotate ASs (see section 2.2 for a detailed description and the GitHub repository for format specifications). The workflow consists of four main steps (see Fig. 1).

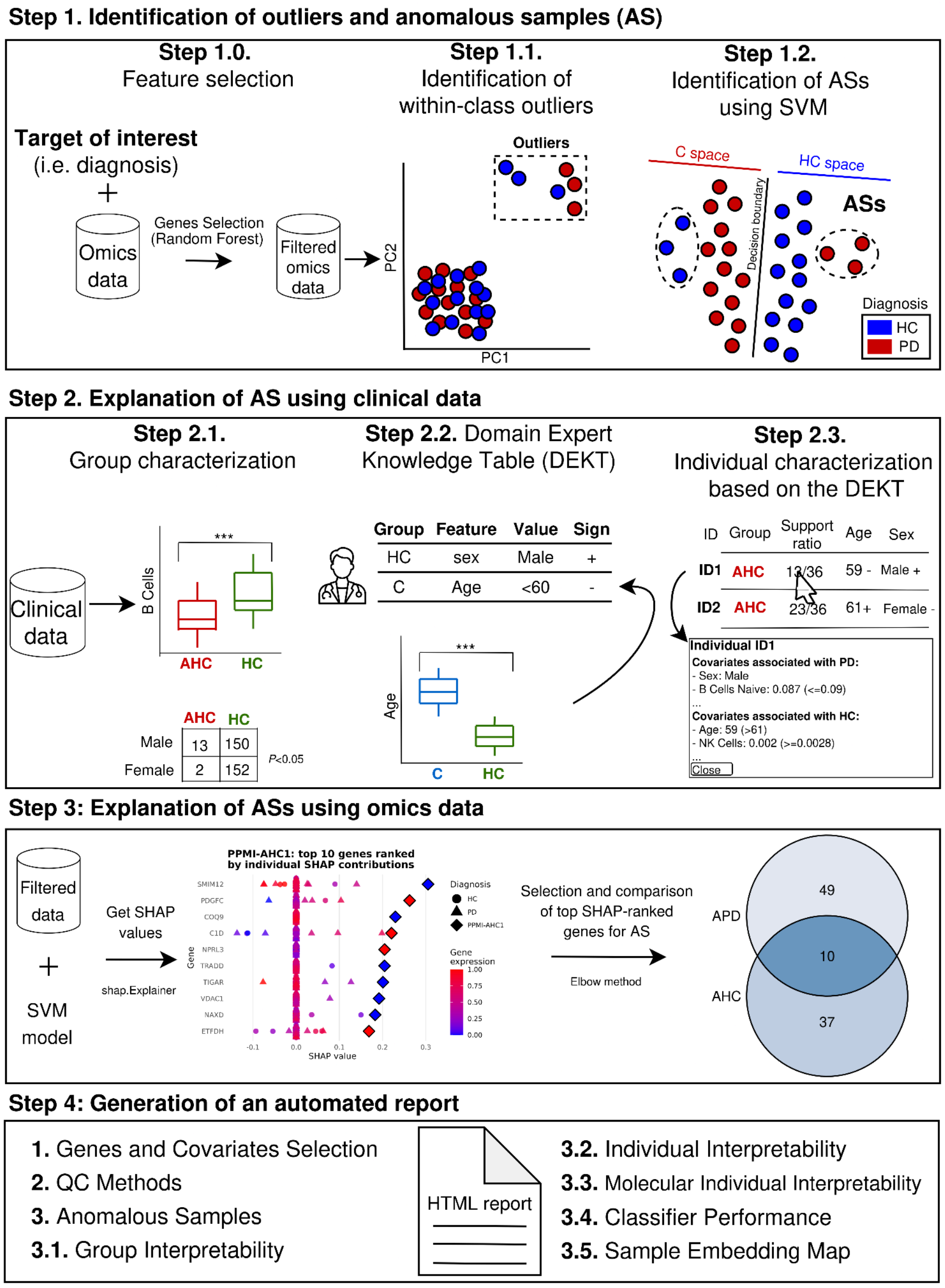

**Fig. 1 MLASDO workflow Step 1: Identification of outliers and ASs. Step 1.0 (Feature selection)**: given the target of interest (i.e., diagnosis), a random forest selects the most relevant genes from the omics data, generating a filtered dataset for downstream analysis. **Step 1.1 (Identification of within-class outliers)**: the filtered data are analysed using two unsupervised methods, Isolation Forest and PCA-Hubert, to identify within-class outliers (dashed boxes). **Step 1.2 (Identification of ASs using SVM)**: an SVM is optimized to find the hyperplane that best separates HCs (blue) and cases (red). Outliers detected by Isolation Forest and PCA-Hubert are labelled accordingly (dashed boxes). For each diagnostic

group, ASs are defined as misclassified samples located beyond a customized threshold from the SVM decision boundary (i.e., HCs falling in the PD space and PD cases falling in the HC space). **Step 2: Explanation of ASs using clinical data. Step 2.1 (Group characterization):** statistical tests identify clinical covariates that distinguish ASs from their original diagnosis group (e.g., AHCs vs. HCs). **Step 2.2 (Disease Expert Knowledge Table, DEKT)**: thresholds and expected directions for each covariate are compiled into the DEKT, which encodes the feature values expected to distinguish the two diagnostic groups. **Step 2.3 (Individual characterization)**: for each AS, the tool generates an interactive table indicating whether each DEKT covariate supports the alternative diagnosis according to its observed value (e.g., a HC with a high proportion of T cells, a feature expected in PD cases). **Step 3: Explanation of ASs using omics data.** SHAP values computed from the SVM models identify the genes contributing to each AS. Individual molecular signatures are then compared to identify the relevant genes shared across ASs within the same diagnostic group. **Step 4: Automated report generation.** MLASDO compiles all results into an interactive report including: (1) the selected genes and clinical covariates together with their functional enrichment; (2) quality-control results, including detected outliers; and (3) AS characterization, including group- and individual-level clinical interpretation, molecular interpretation, classifier performance across SVM kernels, and an embedding map highlighting outliers and ASs within the cohort.

Step 1 starts by selecting the most relevant genes to distinguish between cases and controls (Step 1.0 in Fig. 1). Based on these genes, the tool identifies outliers using unsupervised methods (Step 1.1) and ASs using an SVM (Step 1.2). In Step 2, the tool provides an explanation for ASs based on clinical covariates both at the group and at the individual level. In Step 3, the tool reports, for each AS, the genes with the greatest SHAP contributions to its classification, highlighting those whose expression profiles shift the individual toward the alternative class. Finally, in Step 4, the tool generates an automated interactive report that summarizes the main results of each step of the workflow. It includes a list of samples annotated as ASs, outliers, or both. Specific reports for PPMI and PDBP can be found at the Outputs section of the GitHub repository (https://github.com/joseadrianpp/MLASDO).

## 2.1. Identification of outliers and ASs

This section describes the gene selection procedure, followed by the unsupervised methods used to identify outliers, and the supervised approach for detecting ASs.

### 2.1.0. Gene selection

Genes used in downstream analyses are selected based on their relevance to predict case/control status with a machine learning model (ML). The ML model is created by training a random forest with a 10-fold cross-validation scheme with the ranger and caret R packages [27] and with hyperparameter tuning (see the code on GitHub for details). If class imbalance is present, samples are downsampled within each fold of the training, while validation folds remain unchanged. Additionally, sex and age covariates are added to the pool of candidate variables to select. Feature relevance is quantified via the caret::VarImp function on the best random forest obtained. The final number of selected features is determined by applying a knee-point detection method to the importance distribution using the uik function from the inflection R package [28].

### 2.1.1. Outlier detection

Based on selected genes, outliers are identified by using two unsupervised complementary detectors: Isolation Forest and robust PCA-Hubert. Isolation Forest is applied through the isolation.forest function from the isotree R package [29], employing the univariate-split variant (ndim = 1), 100 trees, and a subsampling size of 256 observations per tree. Samples with an anomaly score greater than 0.5 (default) are classified as outliers. Robust PCA-Hubert is used through the function PcaHubert from the rrcov R package [30]. By default, the number of principal components required to explain 85% of the total variance, with a coverage parameter $\alpha$=0.75 and a cutoff distance set at the 0.975 quantile. Outliers are defined as the union of the samples flagged by the two methods. Samples flagged exclusively by the unsupervised approaches are considered poor-quality samples and are recommended for removal.

### 2.1.2. Supervised detection of anomalous samples

To detect ASs, MLASDO trains an SVM classifier to predict case/control status using the relevant features selected as explained above. We recommend training the SVM on a gene expression matrix where counts are log-normalized in the form of $\log_2$ (fragments per million + 1) and only genes with a minimum expression are kept. To train the model, the tool uses the svm linear function from the kernlab R package [31], which supports the linear, radial basis functions, and polynomial kernels. SVM optimization is conducted through a grid-search procedure applied to each kernel (see code on GitHub for details). Model evaluation uses 10-fold cross-validation with the createFolds function from the caret R package. In each fold, 90% of the samples are used to train the model. Downsampling is applied if necessary through the downSample function from the caret R package. The remaining 10% of the samples are used for evaluation purposes and the detection of the AS. Detection is based on the kernel distance to the classifier hyperplane (see Fig. 1, Step 1.2). The further away from the side of the hyperplane corresponding to its class label, the more likely it will be selected as ASs (see Online Resource Methods S1 for more details).

## 2.2. Annotation of ASs based on clinical evidences

MLASDO provides an extensive annotation of ASs both at the group and individual levels (see Fig. 1, Step 2). At the group level, the tool looks for clinical differences between the whole AS group and the samples from the same diagnostic group, i.e., AHC versus HC and APD (anomalous PD) versus PD. Each AS is also annotated individually based on thresholds for each covariate. These thresholds can be either manually set by an expert clinician or automatically generated based on the dataset. All these thresholds appear in the DEKT, which is specific to each CCC as it depends on the phenotype and the covariates available. The DEKT is used for automated annotation of sample groups and individuals.

### 2.2.1. Group annotation of ASs

A group annotation for ASs is an automatically generated statement describing a covariate whose average value differs significantly between an AS group (i.e., AHCs or anomalous cases) and its corresponding reference group, thereby helping explain why these samples are considered anomalous. The annotation method depends on the type of covariate, including whether it is numeric or categorical, and whether it is ordinal.

For each numeric covariate of interest, MLASDO compares the AS group's distribution with that of their reference group (either controls or cases) through a Mann-Whitney U test and reports the p-value if significant (see Fig. 1, Step 2.1). This test is applied through the wilcox.test function of the stats R package. The resulting statement is a new annotation added to the group's annotation set and additionally illustrated with a whisker plot that displays the distribution of values for both the anomalous and the reference (see section 3.2).

For categorical binary covariates, a contingency table is generated showing the proportion of ASs and reference group samples across the two levels of the covariate. If there is a significant association between levels and belonging to the anomalous and normal group, a new annotation in the form of the contingency table is generated for that covariate. For categorical unordered variables of three or more levels, the proportion of a specific level is compared to the other possible values within each group. For ordinal factors with an arbitrary large number of levels (e.g., the Epworth Sleepiness Scale in the case of Parkinson's Disease), levels are divided into two groups to separate typical values for HCs from those for cases, through a threshold set by the expert clinician. A contingency table is then created, contrasting HCs and cases against individuals over and under the threshold. When at least one of the contingency table cells has a value under 5, a Fisher's exact test is used and a Chi-square test otherwise, implemented through fisher.test and chisq.test functions, respectively, from the stats R package. This manually set threshold is incorporated into the DEKT table and is also useful to annotate individual samples (see below).

### 2.2.2. Individual annotation of ASs

For each AS, each annotation indicates whether the individual's value for that covariate is consistent with its reference group. Covariates with values more characteristic of the alternative group are classified as supporting the change, whereas the remaining covariates are considered normal annotations (Fig. 1, Step 2.3). Together, these annotations provide evidence supporting or refuting the anomalous status of the individual. For example, if the individual is an AHC, the reference group is HC. Let us suppose the individual shows a high proportion of circulating T-cells. This covariate will be listed among the variables that support the change for this individual, since high values of T-cells are associated with a PD-like phenotype. This covariate is classified as supporting the change because it shifts the individual toward the PD phenotype. The higher the number of covariates that support the change received by an individual (i.e., summarized as its support count), the stronger is the evidence supporting its change.

We already noted that these annotations require the appropriate threshold that may be set up by an expert clinician. Alternatively, this generation can be automated. In that case, for numerical covariates, the tool statistically tests for significant differences in mean values between the two clinical groups (e.g., HCs and PD) with a Mann-Whitney U test (wilcox.test R function). If significant, then this variable is of interest for individual annotation. And the corresponding entry is added to the DEKT using as threshold the mean of medians between the two groups (see section 3.3 for examples). For categorical covariates, the procedure is identical to the one used to build the contingency tables as described above. For this paper, entries for the DEKT in regard to PPMI were created by the clinical experts amongst the authors based on the clinical features of the disease (see the corresponding reports for PPMI and PDBP as generated for the DEKT and all annotation results). The report includes an interactive table where the user can inspect, for each individual, the corresponding clinical covariates supporting the individual as anomalous and as normal. This functionality is implemented using the datatable and JS functions from the DT [32] and htmlwidgets [33] R packages, respectively.

## 2.3. Omics-based annotation of ASs

MLASDO provides an omic signature for each AS individually (see Fig. 1, step 3), i.e., the list of genes that were relevant within the SVM model that pointed to it as anomalous. To achieve this, it applies SHAP to each SVM model of the cross-validation where ASs were detected. The tool uses the SHAP Python library, integrated into R through the reticulate package. For details on how the tool manages specific SHAP explainers, see the Online Resource Methods S2. Specifically, for an AHC, the tool detects the genes that support the new label (i.e., PD case instead of HC) by selecting those with positive SHAP values. For APD, the genes with negative SHAP values are the ones that support the new label of these individuals (i.e., HC instead of PD case). Finally, the tool evaluates the presence of a shared transcriptomic signature among anomalous individuals of the same group by identifying the intersection of genes selected for those individuals. These molecules are further annotated through a functional enrichment analysis by using the gost function from the gprofiler2 [34] R package. To this end, the Gene Ontology (GO) [35], Kyoto Encyclopedia of Genes and Genomes (KEGG) [36], REACTOME (REAC) [37], and Human Phenotype (HP) [38] databases are used, discarding *in-silico* annotations and using gSCS to correct for multiple testing.

## 2.4. Structure of the generated report

MLASDO generates an interactive report for each cohort used as input. It gathers the outputs of all steps of the workflow, facilitating interpretability and reproducibility (see Fig. 1, step 4 and the PPMI and PDBP reports available at GitHub). It uses a fixed structure of three sections. Section 1 is dedicated to selecting relevant genes and covariates. In this section, Fig. 1.1 in the report shows how feature importance evolves, displaying genes and covariates in decreasing order of importance. Table 1.1 of the report lists genes and covariates and includes a button for downloading them as an excel file. Fig. 1.2 shows the enrichment of this set of genes. Section 2 of the report presents the summary of results of the unsupervised outlier detection methods in Table 2.1 and the list of samples detected that the user can obtain as an excel file in Table 2.2. The third section of the report provides clinical interpretability of the ASs at the group level (Section 3.1), individual level (Section 3.2), omics level (Section 3.3), SVM performance when classifying samples using the omics (Section 3.4), and an embedding space showing the outliers and ASs with the rest of the samples (Section 3.5). It includes performance of all evaluated SVM kernels, reporting metrics and optimal hyperparameters for each kernel, and explicitly identifying the best-performing model.

### 2.5. Cohort of study

To showcase MLASDO, we used the PPMI cohort, which includes longitudinal clinical data and blood RNA-seq transcriptomics of PD patients and controls. We included 473 individuals at the baseline visit, including 309 idiopathic PD cases (average age of 63 years, men/women ratio=1.88) and 164 HC (average age of 60 years, men/women ratio=1.83), and removed all genetic cases, i.e., those with known mutations in PD-relevant genes (*GBA1*, *SNCA*, *LRRK2*). The list of enrolment criteria applied by PPMI to define idiopathic PD cases is detailed in the Online Resource Methods S3. PPMI includes a large list of clinical covariates, including motor and non-motor information, cell blood counts, imaging data such as specific binding ratio (SBR), concentration of CSF proteins (β-amyloid, Tau, and p-Tau), and patient independence metrics (see Online Resource Table S1). Additionally, baseline transcriptomic data were available for all individuals. Gene expression was corrected for the clinical and technical covariates explaining the highest proportion of transcriptomic variance, namely plate, mature neutrophil proportion, percentage of usable bases, sex, and naïve B-cell proportion. Correction was performed on log-transformed expression values ($\log_2$(FPM + 1)) using the removeBatchEffect function from the *limma* R package [39]. After correction, genes with low expression were removed by retaining only those with an expression value >8 in at least 80% of samples, reducing the dataset from 58,780 to 18,686 genes. Finally, only protein-coding genes, identified through the getBM function of the biomaRt [40] R package, were retained for downstream analysis, leading to a final set of 12,910 genes. Details about the additional cohort studied, PDBP, appear in the Online Resource Methods S4.

## 3. Results

The following sections illustrate the applicability of MLASDO on the PPMI cohort. Description results for the PDBP cohort appear in the Online Resource Results S1. The automatic report generated for each cohort is available on GitHub (https://github.com/joseadrianpp/MLASDO). The tool requires as input (1) a matrix of gene readouts (i.e., RNA expression, proteomics, or methylation are acceptable, provided that features refer to gene loci) with samples in rows and loci in columns and (2) a file with the table containing the clinical/demographical data for those samples, with the same disposition in rows and columns. Samples in both tables must be linked through a shared identifier. Besides, the second table should contain a column with each sample condition. An additional optional input is the DEKT containing, for each row, covariates and their threshold values to separate conditions.

### 3.1. Labelling of outliers and ASs based on omics data

In the PPMI cohort, MLASDO preselected 1,613 genes and age at baseline as relevant features to be used in downstream analyses (see Fig. 1.1 and Table 1.1 in the PPMI report generated by the tool). Particularly, these genes were involved in mitochondria-related pathways, including mitochondrion (GO:0005739, $P<4.93.10^{-8}$), oxidative phosphorylation (KEGG:00190, $P<4.74.10^{-5}$), and respiratory chain complex ($P<1.14.10^{-3}$, GO:0098803) terms (see Fig. 1.2 of the PPMI report). Based on these preselected genes, MLASDO flagged 26 outliers, 11 by Isolation Forest alone; 8 by PCA-Hubert and 7 by both (see Tables 2.1 and 2.2 at the PPMI report). Samples identified exclusively as outliers were flagged for removal from downstream analyses. In contrast, samples identified as ASs, either alone or simultaneously classified as outliers, were retained for further characterization. This resulted in a final set of 5 AHCs and 7 APD cases (see Supplementary Fig. S1, section 3 of the report). The performance of the SVM classifiers with linear, polynomial, and radial kernels is reported in Online Resource Table S2.

## 3.2. Detection of common clinical patterns in ASs

Following the identification of the ASs (i.e., AHCs and APDs), MLASDO proceeded to characterise them at the group level. In order to do so, the clinical profiles of each AS group were compared against those of their original diagnostic group (i.e., AHC vs. HC and APD vs. PD). Among AHCs, despite their absence of a PD diagnosis, they exhibited transcriptomic profiles classified on the PD side of the decision boundary. We therefore investigated whether these molecular anomalies were associated with distinctive clinical features. Specifically, AHCs exhibited substantially higher CSF $A\beta_{1-42}$ levels than HCs (1415.75 [931.88, 1899.62] 95% CI and 1005.39 [922.79, 1087.99] 95% CI pg/mL on average, respectively, $P<0.0396$) (see Fig. 2a). Since lower CSF $A\beta_{1-42}$ levels have been associated with faster cognitive decline in PD [41], these elevated levels suggest that AHCs may represent a subgroup with increased resilience to cognitive decline. In addition, AHCs showed a significantly higher prevalence of severely depressed mood (UPDRS Part I item 2.3 score ≥3 vs. <3, $P<0.0307$;) and anxious mood (UPDRS Part I item 2.4 score ≥3 vs. <3, $P<0.031$; Fig. 2b-c). Anxiety and depression are among the most common non-motor manifestations preceding the onset of motor PD, often emerging even 10 years before clinical diagnosis [42]. Together, these findings suggest that AHCs represent a biologically distinct subgroup that would likely remain undetected using conventional clinical assessments alone, highlighting the potential of transcriptomic profiling to identify clinically relevant molecular heterogeneity before diagnosis.

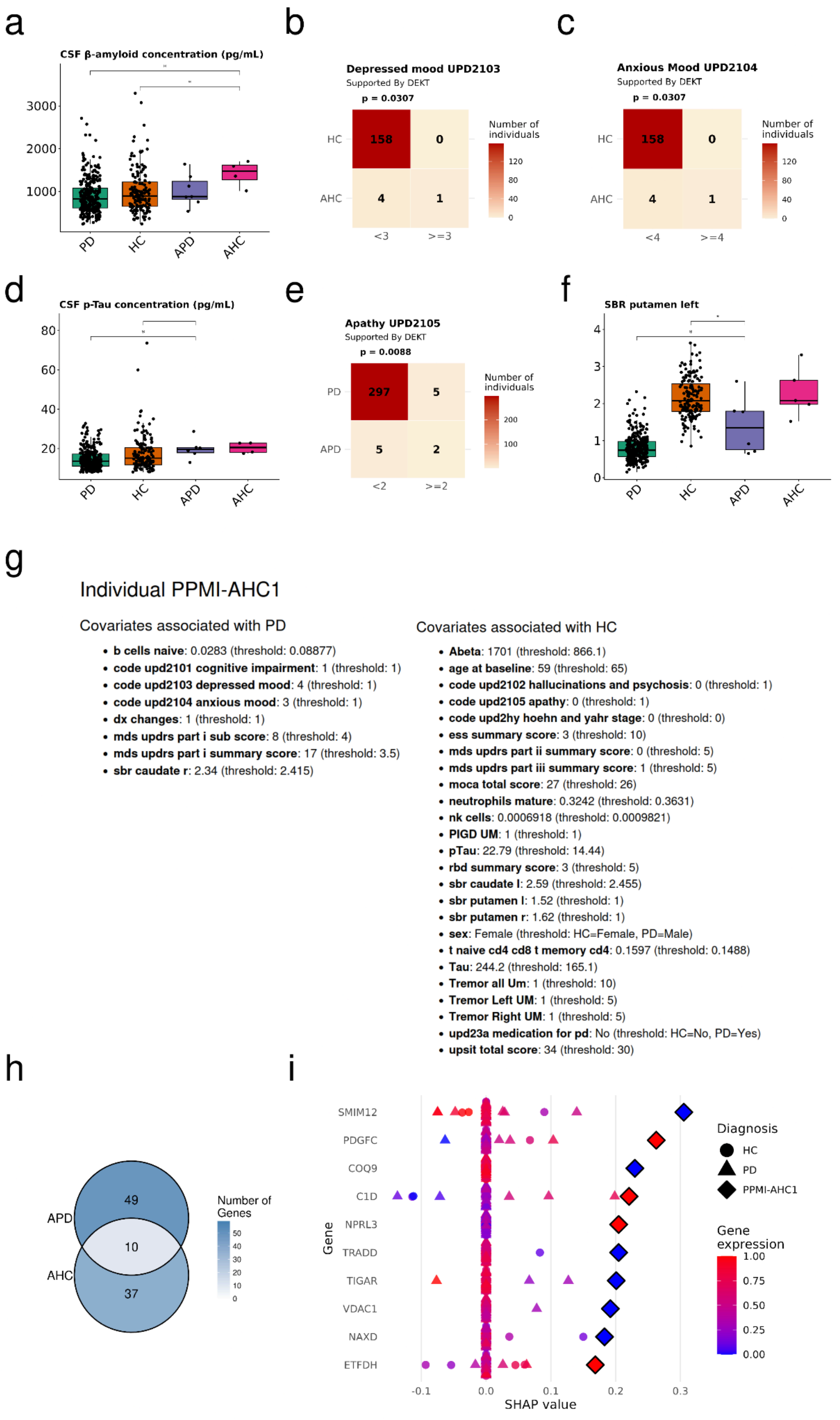

**Fig. 2 Identification and characterization of ASs in the PPMI cohort** The figure shows the outputs generated by MLASDO for the characterization of ASs in the PPMI cohort. The first two rows present the group-level clinical characterization of AHCs (a–c) and APDs (d–f). The third row shows an example of the individual clinical report generated for PPMI-AHC$_1$ (g), while the fourth row presents the group- and individual-level molecular characterization (h–i). **a)** Distribution of CSF β-amyloid concentrations across HC, AHC, PD, and APD groups. Pairwise comparisons between AHCs and HCs were performed using the Mann–Whitney test (* $P < 0.05$, ** $P < 0.01$, *** $P < 0.001$). To facilitate interpretation,

MLASDO annotates each plot according to the DEKT: **Supported by DEKT** indicates that the observed pattern agrees with the expected clinical behaviour, **Not Supported by DEKT** indicates disagreement, and **Unknown by DEKT** indicates that no DEKT rule is available for that covariate. **b–c)** Contingency tables showing the proportion of AHCs and HCs above and below clinically relevant thresholds for severe depressed mood **(b)** and moderate anxious mood **(c).** *P*-values were calculated using Fisher's exact test. **d)** Distribution of CSF p-Tau concentrations across the HC, AHC, PD and APD groups. **e)** Contingency table showing the proportion of APD and PD cases with moderate apathy. **f)** Left putamen specific binding ratio (SBR) (e) across HC, AHC, PD, and APD groups. **g)** Pop-up window from the MLASDO report for PPMI-$AHC_1$, listing the clinical features associated with a PD-like phenotype and those supporting a healthy phenotype, together with the corresponding DEKT thresholds. **h)** Venn diagram showing the overlap between the relevant genes identified for AHCs and APDs. The gene sets correspond to the union of the relevant genes identified across all AHCs and all APDs, respectively. **i)** SHAP beeswarm plot of the most relevant genes associated with the PD-like molecular profile of PPMI-$AHC_1$. Each row corresponds to one selected gene, and each point represents a sample from the PPMI cohort. Point shape indicates the sample group (HC, circle; PD, triangle; PPMI-$AHC_1$, diamond), while colour represents the expression level of the corresponding gene (blue, low; red, high). The x-axis shows the SHAP value, which quantifies the contribution of each gene to the model prediction. For each selected gene, the SHAP value of PPMI-$AHC_1$ is markedly different from those of the remaining samples, highlighting the genes that most strongly contribute to its PD-like molecular profile.

Instead, APD individuals, despite being classified on the healthy side of the transcriptomic decision boundary, exhibited several clinical features consistent with a distinct PD subtype rather than a milder disease phenotype. For example, APD individuals showed higher levels of CSF p-Tau compared to PD cases (19.826 [14.413, 25.239] CI 95% and 14.865 [14.234, 15.497] pg/mL on average, respectively, $P<0.019$, Fig. 2d). Previous studies support this observation. In this line, Bissacco *et al.* reported that *de novo* PD patients with elevated CSF p-Tau levels are at increased risk of developing motor complications, including motor fluctuations [43]. Similarly, Xu *et al.* showed that PD individuals with higher baseline concentrations of CSF p-$Tau_{181}$ were associated with a higher increase in MDS UPDRS Part III score over time, indicating faster motor deterioration [44]. Consistent with these findings, APD individuals exhibited greater motor deterioration during the first year of follow-up compared with PD (MDS-UPDRS Part III $\Delta(M_{12}\text{-}M_0)$ of 6.833 and 3.427, respectively, for APD cases (n=6) and PD cases (n=241)). Moreover, APDs exhibited a significantly higher proportion of individuals with moderate apathy relative to those with low apathy scores than the overall PD cohort (score ≥2 vs score <2, $P<0.0088$, see Fig. 2e). Specifically, both apathy and anxiety have been identified as important predictors of longitudinal severity of motor fluctuations and dyskinesias in patients with *de novo* PD [45], further supporting this APD group as a clinically meaningful molecular subtype with a more complicated motor course. Interestingly, despite this more aggressive clinical profile, APDs also exhibited significantly higher dopamine transporter binding in the left putamen than the remaining PD cohort, as measured by SPECT imaging (1.408 [0.592, 2.224] 95% CI and 0.811 [0.772, 0.851] 95% CI SBR on average, respectively, $P<0.043$) (see Fig. 2f). Although APDs were classified on the healthy side of the transcriptomic decision boundary, they consistently exhibited clinical features associated with faster disease progression. This apparent discrepancy indicates that the transcriptomic features maximizing PD-HC discrimination are not necessarily those driving disease progression. Rather than representing clinically milder patients, APDs appear to follow an alternative molecular trajectory that is underrepresented in the dominant PD transcriptomic signature learned by the SVM.

## 3.3. Individual characterization of ASs

To characterize each AS individually (see Methods and results in section 3.2. of the report), our tool generates a threshold for each numerical covariate by comparing the mean between HCs and PD cases. These thresholds serve as the basis for the DEKT table entries. Alternatively, the tool allows users to manually specify thresholds for any covariate if they prefer to use custom values. For PPMI, we customized some of them according to previous literature. For example, for MOCA we used a clinically established threshold of 26 following the original instrument recommendation [46]. See Online Resource Table S1 for the specific DEKT definition used for PPMI (Table 3.2.2 in the report). Individual results for each AS are gathered in Table 3.2.1 of the report and Online Resource Table S3. Column *support_count* indicates, for the corresponding AS, either AHC or APD, the events supporting their signaling as anomalous, i.e., the change of group for that sample. Higher values in *support_count* point towards more types of evidence supporting the change. By clicking on each individual in Table 3.2.1, one can see a detailed description of all events for that individual.

See, for example, individual PPMI-$AHC_1$ in Table 3.2.1 of the report (also in Fig. 2g): female, age 59, shows a low level of dopaminergic transporter in the right area of the caudate (SBR of 2.34), expected in PD cases (values under 2.415 are associated with a PD-like phenotype). Moreover, it scored high at the MDS-UPDRS Part I test, designed to evaluate non-motor experiences of daily living (observed value 17, while values over 4 are associated with a PD-like phenotype). Besides, this individual showed high values for depressed mood (4/4 score) and anxious mood (3/4 score) but no motor symptom whatsoever (MDS-UPDRS Part III summary score=1). Three months after enrollment in the study -please remind that this analysis is only for baseline data- the participant received a diagnosis of another neurological disorder rather than PD. In this case, the tool might have helped signal this individual at baseline to apply an appropriate follow-up.

## 3.4. Omic signature of ASs at the individual level

For the AHC group, our tool detected 47 genes that are relevant for the classifier to identify samples within the group (see Fig. 2h, section 3.3 in the PPMI report and methods). Notably, as the composition of the AHC group is rather heterogeneous, we expect little functional enrichment for the whole group. Still, these genes were weakly enriched in *Metabolism of vitamins and cofactors* (REAC:R-HSA-196854, $P<0.0267$), including relevant genes such as *PDZD11, COQ9, NAXD, CYB5A,* and *FOLR2,* and in the MAPK signaling pathway (KEGG:04010, $P<0.0145$), implicating genes such as *CACNA2D4, PDGFC, TRADD, DUSP1,* and *MRAS*. While 78.72% (37/47) of the genes were exclusively relevant for AHC, 10 genes were also relevant for some APD individuals with a significant overlap (Chi-squared test, $P<4.78.10^{-16}$). Notably, all five *MAPK* genes fall within the AHC-exclusive subset. In this line, [47] Wang *et al.* demonstrated that phosphorylated p38MAPK and phosphorylated JNK levels were significantly increased in PD cases. The complete list of relevant genes selected for each AS of the PPMI cohort is provided in Online Resource Table S4.

Going back to individual PPMI-$AHC_1$, SHAP detects 10 genes of relevance for it (see Fig. 2i and Table 3.3.1 at the report). Interestingly, some of these genes have already been linked to PD pathology. The *VDAC1* gene, voltage-dependent anion channel 1, encodes a protein that forms a channel through the mitochondrial outer membrane and acts as a critical substrate of *PARKIN*, responsible for the regulation of mitophagy and apoptosis [48]. Chu *et al.* [49] demonstrated that relative levels of *VDAC1* expression were significantly decreased in PD neurons when compared to age-matched controls, and this decrease was significantly greater in nigral neurons with α-synuclein inclusions. Moreover, *TIGAR*, the TP53-induced glycolysis and apoptosis regulator, regulates the pentose phosphate pathway by inhibiting glycolysis and protects against oxidative stress, reduces inflammation, maintains mitochondrial function, and inhibits apoptosis [50]. *TIGAR* is downregulated in PD cases because this protein is specific to Lewy bodies [51]. Regarding the *PDGFC* gene, platelet-derived growth factor C, genes from the *PDGF* family appear upregulated in plasma in Lewy body disease patients with cognitive fluctuations and visual hallucinations [52].

# 4. Discussion

We showcased MLASDO on the PPMI and PDBP cohorts, focusing on PPMI in the main text. Both cohorts include longitudinal clinical data that allow the assessment of whether ASs exhibit faster disease progression than the rest of the cohort. In PPMI, the tool detected five AHC samples whose transcriptomics resemble those of PD cases. A healthy 59-year-old female with strong non-motor experiences in daily living (MDS-UPDRS part I summary score of 17) was diagnosed three months later with an alternative neurological disorder. Interestingly, relevant genes selected for PPMI-$AHC_1$ were already defined as potential therapeutic targets for PD. For example, inhibition of *VDAC1* expression, in both rats and hippocampal neuronal HT22 cells overexpressing α-syn, has been shown to prevent mitochondrial fragmentation and protect dopaminergic neurons [53]. Additionally, Feng *et al.* [54] demonstrated that administration of resveratrol resulted in a significant decrease in the protein expression of *VDAC1* and in the protein expression of α-syn, both *in vivo* and *in vitro*. As another example, Sheng *et al.* [55] demonstrated that *TIGAR* overexpression markedly increased dopaminergic neurons and reduced iron in an MPTP-induced *in vivo* PD model. These results suggest that the use of transcriptomics data is suitable for the early detection of ASs that will possibly develop PD or other neurological disorders and, therefore, are of interest to be followed up over time.

Although omics data has already been used for personalized medicine purposes, these previous attempts were focused on the identification of patient subgroups using unsupervised methods [56,57]. Our tool, in contrast, identifies samples with a supervised approach by contrasting patient groups with what is supported by transcriptomics. Subsequently, it annotates and explains why those samples are anomalous on the basis of their clinical observations.

Our tool individually annotates ASs to characterize them. This characterisation is explained in the report using the DEKT as the basis to produce those explanations. This DEKT is specific to each phenotype and cohort. Generating a new DEKT for a different disease requires an expert to figure out appropriate thresholds for critical clinical covariates. Although the tool can automatically generate the whole DEKT, thresholds for all of the critical variables should be revised by an expert. In addition, the tool generates an interactive table for each AS that shows the clinical covariates that associate this AS with (1) the alternative diagnosis group (i.e., detected using omics data). This allows the user to manually explore and verify each of these features. Notably, although DEKT customization shifted some individual annotations (see section 4.3), the overall ability of the DEKT to label covariates supporting the alternative diagnosis was preserved, so users without a curated DEKT still obtain informative annotations from the automatically generated one.

Finally, we are aware that technical effects and other covariates (e.g., cell type concentrations) can influence transcriptomics, resulting in outliers and ASs that are solely the consequence of these effects. To avoid this, we assess the variance explained by each variable prior to training the SVM and correct it if necessary. Additionally, the most important variables are selected while controlling for age and gender.

# 5. Conclusion

We have presented MLASDO, an open-source and easy to use tool that identifies outliers and anomalous samples (ASs) from transcriptomics data and explains why each AS differs from its original diagnostic group. A single function call is enough to (1) detect outliers and ASs, (2) characterize them at the clinical level, (3) characterize ASs at the molecular level and (4) generate an automated, interactive report, organised into numbered sections with descriptive captions to help the user with the interpretation of the results. The tool's GitHub repository includes a detailed tutorial on how to use it. The objective of MLASDO is not to replace expert clinical judgement; rather, it is designed to support it. MLASDO does this by systematically labelling and explaining discrepancies between clinical diagnosis and molecular profile. MLASDO thus guides researchers towards individuals who most warrant closer inspection. In this way, MLASDO helps build cleaner, better-characterised cohorts for downstream disease research.

# Information Sharing Statement

This study uses clinical and transcriptomics data from the PPMI (NCT01141023) and PDBP (NCT01767818) PD cohorts, requested through their respective official data-access repositories. MLASDO v1.0.0, the software tool developed in this study, is open-access and publicly available at https://github.com/joseadrianpp/MLASDO, with the v1.0.0 release permanently archived on Zenodo at https://zenodo.org/records/22109725. The repository includes a tutorial for running MLASDO on new datasets and the automatically generated anonymous reports for the PPMI and PDBP cohorts.

## Acknowledgements

PPMI – a public-private partnership – is funded by the Michael J. Fox Foundation for Parkinson's Research and funding partners, including 4D Pharma, Abbvie, AcureX, Allergan, Amathus Therapeutics, Aligning Science Across Parkinson's, AskBio, Avid Radiopharmaceuticals, BIAL, BioArctic, Biogen, Biohaven, BioLegend, BlueRock Therapeutics, Bristol-Myers Squibb, Calico Labs, Capsida Biotherapeutics, Celgene, Cerevel Therapeutics, Coave Therapeutics, DaCapo Brainscience, Denali, Edmond J. Safra Foundation, Eli Lilly, Gain Therapeutics, GE HealthCare, Genentech, GSK, Golub Capital, Handl Therapeutics, Insitro, Jazz Pharmaceuticals, Johnson & Johnson Innovative Medicine, Lundbeck, Merck, Meso Scale Discovery, Mission Therapeutics, Neurocrine Biosciences, Neuron23, Neuropore, Pfizer, Piramal, Prevail Therapeutics, Roche, Sanofi, Servier, Sun Pharma Advanced Research Company, Takeda, Teva, UCB, Vanqua Bio, Verily, Voyager Therapeutics, the Weston Family Foundation and Yumanity Therapeutics.

Data and biospecimens used in preparation of this manuscript were obtained from the Parkinson's Disease Biomarkers Program (PDBP) Consortium, supported by the National Institute of Neurological Disorders and Stroke at the National Institutes of Health. Investigators include: Roger Albin, Roy Alcalay, Alberto Ascherio, Thomas Beach, Sarah Berman, Bradley Boeve, F. DuBois Bowman, Shu Chen, Alice Chen-Plotkin, William Dauer, Ted Dawson, Paula Desplats, Richard Dewey, Ray Dorsey, Jori Fleisher, Kirk Frey, Douglas Galasko, James Galvin, Dwight German, Steven Gunzler, Lawrence Honig, Xuemei Huang, David Irwin, Un Kang, Kejal Kantarci, Anumantha Kanthasamy, Daniel Kaufer, Horacio Kaufmann, Qingzhong Kong, James Leverenz, Allan Levey, Carol Lippa, Irene Litvan, Oscar Lopez, Jian Ma, Richard Mailman, Lara Mangravite, Karen Marder, Kelly Mills, Nandakumar Narayanan, Laurie Orzelius, Vladislav Petyuk, Judith Potashkin, Liana Rosenthal, Rachel Saunders-Pullman, Clemens Scherzer, Michael Schwarzschild, Nicholas Seyfried, Tanya Simuni, Andrew Singleton, David Standaert, Debby Tsuang, David Vaillancourt, Jerrold Vitek, David Walt, Andrew West, Cyrus Zabetian, and Jing Zhang. The PDBP Investigators have not participated in reviewing the data analysis or content of the manuscript.

## Authors' contributions

**José Adrián Pardo-Pérez:** Formal analysis, Software, Visualization, Writing – original draft. **Jaime Ñíguez:** Data curation. **Ana Luisa Gil Martínez:** Supervision, Validation. **Tomás Bernal:** Formal analysis, Software. **Guillermo Rocamora:** Data curation. **Mina Ryten:** Resources. **Laura Ibañez:** Writing - review and editing. **José T. Palma:** Writing - review and editing. **Juan A. Botía:** Conceptualization, Methodology, Supervision. **Alicia Gómez-Pascual:** Supervision, Validation, Writing – original draft. **All authors:** Writing – review and editing.

## Funding

JAPP was funded by Fundación Séneca. Región de Murcia (Spain) [Project number: 22801/FPI/24]. JÑ was funded by the Spanish Ministry of Science, Innovation and Universities (MCIN/AEI/10.13039/501100011033) and the European Social Fund Plus (ESF+) [Predoctoral grant number: PREP2022-000279; Project number: PID2022-136306OB-I00]. ALGM was funded by Fundación Séneca. Región de Murcia (Spain) [Project number: 22821/SF/24]. AGP was supported by RIAM INTELEARNING LAB, S.L. through the IARECGNN project.

## Data availability

Transcriptomics data from the PPMI cohort were requested from the official website (https://www.ppmi-info.org/access-data-specimens/download-data) (RRID:SCR_006431). Transcriptomics data from the PDBP cohort were requested through the PDBP Data Management Resource (https://pdbp.ninds.nih.gov/data-management-resource).

## Code availability

MLASDO v1.0.0 is an open-access software available for the scientific community at: https://github.com/joseadrianpp/MLASDO. In this repository, you can find both a tutorial on how to run MLASDO on your own dataset and the automatic reports that MLASDO generated for PPMI and PDBP cohorts.

## Statements and Declarations

### Ethics approval and consent to participate

This study involves human samples obtained from PPMI (NCT01141023) and PDBP cohorts (NCT01767818). All participants provided written informed consent at the time of enrollment in the original studies in accordance with the ethical standards of the institutional and national research committee and with the 1964 Helsinki declaration and its later amendments or comparable ethical standards. The study protocol and informed consent forms of PPMI are available at the PPMI website (20/LO/0900, 7 Sep 2020). The consent form of PDBP clearly states that the collected data would be stored in the PDBP database and de-identified data shared with PDBP Consortium researchers and/or other investigators.

### Consent for publication

This article does not contain any data from which individual participants can be identified.

### Competing Interests

The authors declare no competing interests.

# References

[1] D.J. Hunter, Gene–environment interactions in human diseases, Nat Rev Genet 6 (2005) 287–298. https://doi.org/10.1038/nrg1578.

[2] D. Berg, P. Borghammer, S.-M. Fereshtehnejad, S. Heinzel, J. Horsager, E. Schaeffer, R.B. Postuma, Prodromal Parkinson disease subtypes — key to understanding heterogeneity, Nat Rev Neurol 17 (2021) 349–361. https://doi.org/10.1038/s41582-021-00486-9.

[3] D. Ferreira, A. Nordberg, E. Westman, Biological subtypes of Alzheimer disease, Neurology 94 (2020) 436–448. https://doi.org/10.1212/WNL.0000000000009058.

[4] Parkinson Progression Marker Initiative, The Parkinson Progression Marker Initiative (PPMI), Prog Neurobiol 95 (2011) 629–635. https://doi.org/10.1016/j.pneurobio.2011.09.005.

[5] L.S. Rosenthal, D. Drake, R.N. Alcalay, D. Babcock, F.D. Bowman, A. Chen-Plotkin, T.M. Dawson, R.B. Dewey, D.C. German, X. Huang, B. Landin, M. McAuliffe, V.A. Petyuk, C.R. Scherzer, C.S. Hillaire-Clarke, B.-A. Sieber, M. Sutherland, C. Tarn, A. West, D. Vaillancourt, J. Zhang, K. Gwinn, PDBP consortium, The NINDS Parkinson's disease biomarkers program, Mov Disord 31 (2016) 915–923. https://doi.org/10.1002/mds.26438.

[6] H. Iwaki, H.L. Leonard, M.B. Makarious, M. Bookman, B. Landin, D. Vismer, B. Casey, J.R. Gibbs, D.G. Hernandez, C. Blauwendraat, D. Vitale, Y. Song, D. Kumar, C.L. Dalgard, M. Sadeghi, X. Dong, L. Misquitta, S.W. Scholz, C.R. Scherzer, M.A. Nalls, S. Biswas, A.B. Singleton, Uniformed Services University of the Health Sciences Associates, AMP PD Whole Genome Sequencing Working Group, AMP PD consortium, Accelerating Medicines Partnership: Parkinson's Disease. Genetic Resource, Mov Disord 36 (2021) 1795–1804. https://doi.org/10.1002/mds.28549.

[7] R.C. Petersen, P.S. Aisen, L.A. Beckett, M.C. Donohue, A.C. Gamst, D.J. Harvey, C.R. Jack, W.J. Jagust, L.M. Shaw, A.W. Toga, J.Q. Trojanowski, M.W. Weiner, Alzheimer's Disease Neuroimaging Initiative (ADNI): clinical characterization, Neurology 74 (2010) 201–209. https://doi.org/10.1212/WNL.0b013e3181cb3e25.

[8] C. Bycroft, C. Freeman, D. Petkova, G. Band, L.T. Elliott, K. Sharp, A. Motyer, D. Vukcevic, O. Delaneau, J. O'Connell, A. Cortes, S. Welsh, A. Young, M. Effingham, G. McVean, S. Leslie, N. Allen, P. Donnelly, J. Marchini, The UK Biobank resource with deep phenotyping and genomic data, Nature 562 (2018) 203–209. https://doi.org/10.1038/s41586-018-0579-z.

[9] A.H. Ramirez, L. Sulieman, D.J. Schlueter, A. Halvorson, J. Qian, F. Ratsimbazafy, R. Loperena, K. Mayo, M. Basford, N. Deflaux, K.N. Muthuraman, K. Natarajan, A. Kho, H. Xu, C. Wilkins, H. Anton-Culver, E. Boerwinkle, M. Cicek, C.R. Clark, E. Cohn, L. Ohno-Machado, S.D. Schully, B.K. Ahmedani, M. Argos, R.M. Cronin, C. O'Donnell, M. Fouad, D.B. Goldstein, P. Greenland, S.J. Hebbring, E.W. Karlson, P. Khatri, B. Korf, J.W. Smoller, S. Sodeke, J. Wilbanks, J. Hentges, S. Mockrin, C. Lunt, S.A. Devaney, K. Gebo, J.C. Denny, R.J. Carroll, D. Glazer, P.A. Harris, G. Hripcsak, A. Philippakis, D.M. Roden, All of Us Research Program, The All of Us Research Program: Data quality, utility, and diversity, Patterns (N Y) 3 (2022) 100570. https://doi.org/10.1016/j.patter.2022.100570.

[10] M.M. Breunig, H.-P. Kriegel, R.T. Ng, J. Sander, LOF: identifying density-based local outliers, SIGMOD Rec. 29 (2000) 93–104. https://doi.org/10.1145/335191.335388.

[11] F.T. Liu, K.M. Ting, Z.-H. Zhou, Isolation Forest, in: 2008 Eighth IEEE International Conference on Data Mining, 2008: pp. 413–422. https://doi.org/10.1109/ICDM.2008.17.

[12] B. Schölkopf, A.J. Smola, Learning with Kernels: Support Vector Machines, Regularization, Optimization, and Beyond, The MIT Press, 2001. https://doi.org/10.7551/mitpress/4175.001.0001.
[13] C. Peng, Y. Chen, Z. Kang, C. Chen, Q. Cheng, Robust principal component analysis: A factorization-based approach with linear complexity, Information Sciences 513 (2020) 581–599. https://doi.org/10.1016/j.ins.2019.09.074.
[14] J. Bootkrajang, A. Kabán, Classification of mislabelled microarrays using robust sparse logistic regression, Bioinformatics 29 (2013) 870–877. https://doi.org/10.1093/bioinformatics/btt078.
[15] X. Zhang, X. Lu, Q. Shi, X. Xu, H.E. Leung, L.N. Harris, J.D. Iglehart, A. Miron, J.S. Liu, W.H. Wong, Recursive SVM feature selection and sample classification for mass-spectrometry and microarray data, BMC Bioinformatics 7 (2006) 197. https://doi.org/10.1186/1471-2105-7-197.
[16] M. Baldomero-Naranjo, L.I. Martínez-Merino, A.M. Rodríguez-Chía, A robust SVM-based approach with feature selection and outliers detection for classification problems, Expert Systems with Applications 178 (2021) 115017. https://doi.org/10.1016/j.eswa.2021.115017.
[17] J.P. Brooks, Support Vector Machines with the Ramp Loss and the Hard Margin Loss, Operations Research 59 (2011) 467–479. https://doi.org/10.1287/opre.1100.0854.
[18] J.S. Rhodes, A. Cutler, K.R. Moon, Geometry- and Accuracy-Preserving Random Forest Proximities, IEEE Transactions on Pattern Analysis and Machine Intelligence 45 (2023) 10947–10959. https://doi.org/10.1109/TPAMI.2023.3263774.
[19] Y. Zhao, Z. Nasrullah, Z. Li, PyOD: A Python Toolbox for Scalable Outlier Detection, Journal of Machine Learning Research 20 (2019) 1–7.
[20] N.P.M. Nita-lazar Jian Song,Aleksandra, EnsMOD: A Software Program for Omics Sample Outlier Detection - Nathan P. Manes, Jian Song, Aleksandra Nita-lazar, 2023, Journal of Computational Biology (2023). https://journals.sagepub.com/doi/epdf/10.1089/cmb.2022.0243?cf-mal-redirected=true (accessed May 5, 2026).
[21] J.R. Leary, Y. Xu, A.B. Morrison, C. Jin, E.C. Shen, P.C. Kuhlers, Y. Su, N.U. Rashid, J.J. Yeh, X.L. Peng, Sub-Cluster Identification through Semi-Supervised Optimization of Rare-Cell Silhouettes (SCISSORS) in single-cell RNA-sequencing, Bioinformatics 39 (2023) btad449. https://doi.org/10.1093/bioinformatics/btad449.
[22] A. Jensch, M.B. Lopes, S. Vinga, N. Radde, ROSIE: RObust Sparse ensemble for outlIEr detection and gene selection in cancer omics data, Stat Methods Med Res 31 (2022) 947–958. https://doi.org/10.1177/09622802211072456.
[23] M. Hubert, P.J. Rousseeuw, K. Vanden Branden, ROBPCA: A New Approach to Robust Principal Component Analysis, Technometrics 47 (2005) 64–79. https://doi.org/10.1198/004017004000000563.
[24] C. Cortes, V. Vapnik, Support-vector networks, Mach Learn 20 (1995) 273–297. https://doi.org/10.1007/BF00994018.
[25] S.M. Lundberg, S.-I. Lee, A Unified Approach to Interpreting Model Predictions, in: Advances in Neural Information Processing Systems, Curran Associates, Inc., 2017. https://papers.nips.cc/paper_files/paper/2017/hash/8a20a8621978632d76c43dfd28b67767-Abstract.html (accessed October 7, 2025).
[26] S.M. Lundberg, G. Erion, H. Chen, A. DeGrave, J.M. Prutkin, B. Nair, R. Katz, J. Himmelfarb, N. Bansal, S.-I. Lee, From local explanations to global understanding with explainable AI for trees, Nat Mach Intell 2 (2020) 56–67. https://doi.org/10.1038/s42256-019-0138-9.

[27] M. Kuhn, Building Predictive Models in R Using the caret Package, Journal of Statistical Software 28 (2008) 1–26. https://doi.org/10.18637/jss.v028.i05.
[28] D.T. Christopoulos, inflection: Finds the Inflection Point of a Curve, (2025). https://cran.r-project.org/web/packages/inflection/index.html (accessed January 7, 2026).
[29] D. Cortes, T.G.-P. (Copyright holder of included robinmap library), D.B. (Copyright holder of original xoshiro code), S.V. (Copyright holder of original xoshiro code), N.D. (Copyright holder of formatted ziggurat tables), S.D. (Copyright holder of parts of digamma implementation), E.I. (Copyright holder of parts of digamma implementation), S.M. (Copyright holder of parts of digamma implementation), isotree: Isolation-Based Outlier Detection, (2026). https://cran.r-project.org/web/packages/isotree/index.html (accessed June 19, 2026).
[30] V. Todorov, rrcov: Scalable Robust Estimators with High Breakdown Point, (2025). https://cran.r-project.org/web/packages/rrcov/index.html (accessed June 19, 2026).
[31] A. Karatzoglou, A. Smola, K. Hornik, A. Zeileis, kernlab - An S4 Package for Kernel Methods in R, Journal of Statistical Software 11 (2004) 1–20. https://doi.org/10.18637/jss.v011.i09.
[32] Y. Xie, J. Cheng, X. Tan, J.J. Allaire, M. Girlich, G.F. Ellis, J. Rauh, S.L. (DataTables in htmlwidgets/lib), B.R. (selectize js in htmlwidgets/lib), L.G. (noUiSlider in htmlwidgets/lib), B.S. (jquery highlight js in htmlwidgets/lib), A. Pickering, W. Holmes, M. Marttila, A. Quintero, S. Laurent, P. Software, PBC, DT: A Wrapper of the JavaScript Library "DataTables," (2024). https://cran.r-project.org/web/packages/DT/index.html (accessed December 30, 2024).
[33] R. Vaidyanathan, Y. Xie, J.J. Allaire, J. Cheng, C. Sievert, K. Russell, E. Hughes, P. Software, PBC, htmlwidgets: HTML Widgets for R, (2023). https://cran.r-project.org/web/packages/htmlwidgets/index.html (accessed December 30, 2024).
[34] L. Kolberg, U. Raudvere, I. Kuzmin, J. Vilo, H. Peterson, gprofiler2 -- an R package for gene list functional enrichment analysis and namespace conversion toolset g:Profiler, F1000Res 9 (2020) ELIXIR-709. https://doi.org/10.12688/f1000research.24956.2.
[35] M.A. Harris, J. Clark, A. Ireland, J. Lomax, M. Ashburner, R. Foulger, K. Eilbeck, S. Lewis, B. Marshall, C. Mungall, J. Richter, G.M. Rubin, J.A. Blake, C. Bult, M. Dolan, H. Drabkin, J.T. Eppig, D.P. Hill, L. Ni, M. Ringwald, R. Balakrishnan, J.M. Cherry, K.R. Christie, M.C. Costanzo, S.S. Dwight, S. Engel, D.G. Fisk, J.E. Hirschman, E.L. Hong, R.S. Nash, A. Sethuraman, C.L. Theesfeld, D. Botstein, K. Dolinski, B. Feierbach, T. Berardini, S. Mundodi, S.Y. Rhee, R. Apweiler, D. Barrell, E. Camon, E. Dimmer, V. Lee, R. Chisholm, P. Gaudet, W. Kibbe, R. Kishore, E.M. Schwarz, P. Sternberg, M. Gwinn, L. Hannick, J. Wortman, M. Berriman, V. Wood, N. de la Cruz, P. Tonellato, P. Jaiswal, T. Seigfried, R. White, Gene Ontology Consortium, The Gene Ontology (GO) database and informatics resource, Nucleic Acids Res 32 (2004) D258-261. https://doi.org/10.1093/nar/gkh036.
[36] M. Kanehisa, S. Goto, KEGG: kyoto encyclopedia of genes and genomes, Nucleic Acids Res 28 (2000) 27–30. https://doi.org/10.1093/nar/28.1.27.
[37] L. Matthews, G. Gopinath, M. Gillespie, M. Caudy, D. Croft, B. de Bono, P. Garapati, J. Hemish, H. Hermjakob, B. Jassal, A. Kanapin, S. Lewis, S. Mahajan, B. May, E. Schmidt, I. Vastrik, G. Wu, E. Birney, L. Stein, P. D'Eustachio, Reactome knowledgebase of human biological pathways and processes, Nucleic Acids Res 37 (2009) D619-622. https://doi.org/10.1093/nar/gkn863.

[38] P.N. Robinson, S. Köhler, S. Bauer, D. Seelow, D. Horn, S. Mundlos, The Human Phenotype Ontology: a tool for annotating and analyzing human hereditary disease, Am J Hum Genet 83 (2008) 610–615. https://doi.org/10.1016/j.ajhg.2008.09.017.
[39] M.E. Ritchie, B. Phipson, D. Wu, Y. Hu, C.W. Law, W. Shi, G.K. Smyth, limma powers differential expression analyses for RNA-sequencing and microarray studies, Nucleic Acids Research 43 (2015) e47. https://doi.org/10.1093/nar/gkv007.
[40] S. Durinck, Y. Moreau, A. Kasprzyk, S. Davis, B. De Moor, A. Brazma, W. Huber, BioMart and Bioconductor: a powerful link between biological databases and microarray data analysis, Bioinformatics 21 (2005) 3439–3440. https://doi.org/10.1093/bioinformatics/bti525.
[41] A. Siderowf, S.X. Xie, H. Hurtig, D. Weintraub, J. Duda, A. Chen-Plotkin, L.M. Shaw, V. Van Deerlin, J.Q. Trojanowski, C. Clark, CSF amyloid β 1-42 predicts cognitive decline in Parkinson disease, Neurology 75 (2010) 1055–1061. https://doi.org/10.1212/WNL.0b013e3181f39a78.
[42] F. Bracone, A. Gialluisi, S. Costanzo, M. Bonaccio, T. Panzera, S. Orlandi, M. Persichillo, A. De Curtis, M.I. De Bartolo, D. Belvisi, A. Berardelli, S. Falciglia, C. Cerletti, M.B. Donati, G. de Gaetano, L. Iacoviello, A. Di Castelnuovo, Anxiety and depression as prodromes of Parkinson's disease: prospective findings from the Moli-sani study, J Neurol 273 (2026) 175. https://doi.org/10.1007/s00415-026-13710-7.
[43] J. Bissacco, G. Di Lazzaro, R. Bovenzi, G.M. Sancesario, M. Conti, C. Simonetta, D. Mascioli, M. Mancini, V. Buttarazzi, M. Pierantozzi, M. Pieri, S. Bernardini, A. Stefani, A.R. Bentivoglio, P. Calabresi, N.B. Mercuri, T. Schirinzi, CSF phospho-tau levels at Parkinson's disease onset predict the risk for development of motor complications, J Neurol 272 (2025) 573. https://doi.org/10.1007/s00415-025-13325-4.
[44] F. Xu, H. Huang, J. Feng, Q. Shen, Y. Bao, D. Zhang, Y. Xu, Cerebrospinal fluid tau and disease progression in early Parkinson's disease: an 8-year longitudinal study, J Neurol 272 (2024) 61. https://doi.org/10.1007/s00415-024-12856-6.
[45] J.T. Hinkle, K. Perepezko, L.L. Gonzalez, K.A. Mills, G.M. Pontone, Apathy and Anxiety in De Novo Parkinson's Disease Predict the Severity of Motor Complications, Mov Disord Clin Pract 8 (2021) 76–84. https://doi.org/10.1002/mdc3.13117.
[46] Z.S. Nasreddine, N.A. Phillips, V. Bédirian, S. Charbonneau, V. Whitehead, I. Collin, J.L. Cummings, H. Chertkow, The Montreal Cognitive Assessment, MoCA: a brief screening tool for mild cognitive impairment, J Am Geriatr Soc 53 (2005) 695–699. https://doi.org/10.1111/j.1532-5415.2005.53221.x.
[47] S. Wang, C. Zhang, X. Sheng, X. Zhang, B. Wang, G. Zhang, Peripheral expression of MAPK pathways in Alzheimer's and Parkinson's diseases, J Clin Neurosci 21 (2014) 810–814. https://doi.org/10.1016/j.jocn.2013.08.017.
[48] S. Abu-Hamad, N. Arbel, D. Calo, L. Arzoine, A. Israelson, N. Keinan, R. Ben-Romano, O. Friedman, V. Shoshan-Barmatz, The VDAC1 N-terminus is essential both for apoptosis and the protective effect of anti-apoptotic proteins, J Cell Sci 122 (2009) 1906–1916. https://doi.org/10.1242/jcs.040188.
[49] Y. Chu, J.G. Goldman, L. Kelly, Y. He, T. Waliczek, J.H. Kordower, Abnormal alpha-synuclein reduces nigral voltage-dependent anion channel 1 in sporadic and experimental Parkinson's disease, Neurobiol Dis 69 (2014) 1–14. https://doi.org/10.1016/j.nbd.2014.05.003.
[50] K. Bensaad, A. Tsuruta, M.A. Selak, M.N.C. Vidal, K. Nakano, R. Bartrons, E. Gottlieb, K.H. Vousden, TIGAR, a p53-inducible regulator of glycolysis and apoptosis, Cell 126 (2006) 107–120. https://doi.org/10.1016/j.cell.2006.05.036.
[51] K.L.R. López, J.E. Simpson, L.C. Watson, H. Mortiboys, G.M. Hautbergue, O. Bandmann, J.R. Highley, TIGAR inclusion pathology is specific for Lewy body

diseases, Brain Res 1706 (2019) 218–223. https://doi.org/10.1016/j.brainres.2018.09.032.
[52] P. Youssef, M. Chua, L. Andersson, J. Galper, C.W. Gunawardana, E. Matar, G.M. Halliday, S.J. Lewis, N. Dzamko, Platelet-derived growth factor is increased with cognitive fluctuations and hallucinations in Lewy Body disease, Parkinsonism Relat Disord 141 (2025) 108094. https://doi.org/10.1016/j.parkreldis.2025.108094.
[53] M. Rajendran, M. Queralt-Martín, P.A. Gurnev, W.M. Rosencrans, A. Rovini, D. Jacobs, K. Abrantes, D.P. Hoogerheide, S.M. Bezrukov, T.K. Rostovtseva, Restricting α-synuclein transport into mitochondria by inhibition of α-synuclein–VDAC complexation as a potential therapeutic target for Parkinson's disease treatment, Cell. Mol. Life Sci. 79 (2022) 368. https://doi.org/10.1007/s00018-022-04389-w.
[54] S. Feng, J. Gui, B. Qin, J. Ye, Q. Zhao, A. Guo, M. Sang, X. Sun, Resveratrol Inhibits VDAC1-Mediated Mitochondrial Dysfunction to Mitigate Pathological Progression in Parkinson's Disease Model, Mol Neurobiol 62 (2025) 6636–6654. https://doi.org/10.1007/s12035-024-04234-0.
[55] Y. Sheng, J. Huang, W. Wu, X. Wan, J. Wang, Z. Qin, Y. Wang, TIGAR plays neuroprotective roles in MPP+/MPTP-induced Parkinson's disease by alleviating ferroptosis, European Journal of Pharmacology 995 (2025) 177430. https://doi.org/10.1016/j.ejphar.2025.177430.
[56] X. Huang, A.J. Jannu, Z. Song, N. Jury-Garfe, C.A. Lasagna-Reeves, Alzheimer's Disease Neuroimaging Initiative, T.S. Johnson, K. Huang, J. Zhang, Predicting Alzheimer's disease subtypes and understanding their molecular characteristics in living patients with transcriptomic trajectory profiling, Alzheimers Dement (2025) e14241. https://doi.org/10.1002/alz.14241.
[57] H. Li, M. Wei, T. Ye, Y. Liu, D. Qi, X. Cheng, Identification of the molecular subgroups in Alzheimer's disease by transcriptomic data, Front Neurol 13 (2022) 901179. https://doi.org/10.3389/fneur.2022.901179.

## Author Biographies

**Jose A. Pardo-Pérez** (JPP-UMU), MSc, is a FPI Seneca research fellow at UMU and computer scientist by training. He researches computational models applied to Parkinson's disease.

**Jaime Ñiguez** (JN-UMU), MSc, is a FPI Seneca researcher fellow at UMU and biologist by training. He is working on computational models of Parkinson's disease progression.

**Ana Luisa Gil-Martínez** (ALGM-UMU), PhD, is a Saavedra Fajardo researcher fellow at UMU. She develops methods integrating clinical, genetic and transcriptomic data in neurodegeneration.

**Tomás Bernal** (TB-UMU), MSc, is a FPI Seneca researcher fellow at UMU and computer scientist by training. He is working with Natural Language Processing.

**Guillermo Rocamora** (GR-UKDRI) is researcher at the UK Dementia Research Institute, Cambridge, working with Prof. Ryten on transcriptomic modelling of Parkinson's Disease progression.

**Mina Ryten** (MR-UKDRI), MD, PhD, is a Professor and Director of the UK Dementia Research Institute at Cambridge. She uses human brain transcriptomics to understand neurological diseases.

**Laura Ibáñez** (LI-WUSTL), PhD, is an Assistant Professor at Washington University in St. Louis. She studies RNA-based biomarkers for Alzheimer's and Parkinson's disease diagnosis.

**José T. Palma** (JTP-UMU), PhD, is a computer scientist by training and a machine learning specialist highly experienced in developing artificial intelligence based applications for the medical domain.

**Juan A. Botía** (JAB-UMU), PhD, is the principal investigator in this project and leader of the UMU team. He is a Machine Learning specialist with a focus on bioinformatics of neurological disorders.

**Alicia Gómez-Pascual** (AGP-UMU), PhD, is a postdoctoral researcher with a background in biochemistry, developing computational tools for omics analysis in neurodegenerative diseases.